\documentclass{article} 
\usepackage{iclr2023_conference,times}

\usepackage{amsmath,amsfonts,bm}

\def\eqref#1{equation~\ref{#1}}

\def\1{\bm{1}}

\DeclareMathAlphabet{\mathsfit}{\encodingdefault}{\sfdefault}{m}{sl}
\SetMathAlphabet{\mathsfit}{bold}{\encodingdefault}{\sfdefault}{bx}{n}

\usepackage[margin=1.06in]{geometry}

\usepackage{hyperref}
\usepackage{url}
\usepackage{graphicx} 
\usepackage{subfig} 
\usepackage{multirow} 
\usepackage{booktabs} 
\usepackage{xspace}
\usepackage{array}
\usepackage{color, colortbl}
\usepackage{pifont}

\newcommand{\cmark}{\ding{51}}
\newcommand{\xmark}{\ding{55}}

\title{PopSparse: Accelerated block sparse matrix \\multiplication on IPU}

\author{Zhiyi Li\thanks{equal contriubtion},~~Douglas Orr\textsuperscript{*}, Valeriu Ohan\textsuperscript{*}, Godfrey Da costa, Tom Murray, Adam Sanders, Deniz Beker \& \\ \textbf{Dominic Masters} \\
Graphcore, UK \\
\texttt{\{zhiyil, douglaso, valeriuo\}@graphcore.ai} \\
}

\definecolor{darkgreen}{rgb}{0, 0.6, 0}

\iclrfinalcopy 
\begin{document}

\maketitle
\lhead{}
\begingroup\renewcommand\thefootnote{*}
\renewcommand{\thefootnote}{\arabic{footnote}}
\begin{abstract}
Reducing the computational cost of running large scale neural networks using sparsity has attracted great attention in the deep learning community. While much success has been achieved in reducing FLOP and parameter counts while maintaining acceptable task performance, achieving actual speed improvements has typically been much more difficult, particularly on general purpose accelerators (GPAs) such as NVIDIA GPUs using low precision number formats. In this work we introduce PopSparse, a library that enables fast sparse operations on Graphcore IPUs by leveraging both the unique hardware characteristics of IPUs as well as any block structure defined in the data. We target two different types of sparsity: static, where the sparsity pattern is fixed at compile-time; and dynamic, where it can change each time the model is run. We present benchmark results for matrix multiplication for both of these modes on IPU with a range of block sizes, matrix sizes and densities. Results indicate that the PopSparse implementations are faster than dense matrix multiplications on IPU at a range of sparsity levels with large matrix size and block size. Furthermore, static sparsity in general outperforms dynamic sparsity. While previous work on GPAs has shown speedups only for very high sparsity (typically 99\% and above), the present work demonstrates that our static sparse implementation outperforms equivalent dense calculations in FP16 at lower sparsity (around 90\%). IPU code is available to view and run at \href{http://ipu.dev/sparsity-benchmarks}{\texttt{ipu.dev/sparsity-benchmarks}}, GPU code will be made available shortly.

\end{abstract}

\section{Introduction}\label{Introduction}
The topic of sparsity has gained significant attention in the field of deep learning research due to its potential for increased computational efficiency, reduced model size, and closer alignment with brain-like computation. The notion of sparsity in deep learning most commonly refers to the idea of sparsifying the model weights with the aim of reducing the associated storage and compute costs. This is typically done through a single pruning step at the end of training~\citep{Zhu2017} or potentially through some sparse training regime~\citep{evci2019}. These approaches have typically achieved between 90-99$\%$ reduction in model weights while maintaining and acceptable level of accuracy depending on the model and technique used~\citep{Hoefler2021}.

While the benefits to model size can be easily realised, increased computational efficiency can often be more difficult to achieve as sparse computation can be challenging to execute effectively on modern, highly parallel deep learning accelerators~\citep{Qin2022}. 
As such, sparse training methods often fall into a gap between theoretical and practical efficiency.
For example \citet{Mostafa2019} demonstrated FLOP efficient deep residual CNN training with dynamic sparse reparameterisation techniques but struggled to achieve speed ups in practice. 
A key reason behind this is that unstructured sparsity patterns do not align well with the vectorised instruction sets that have been so successful in accelerating highly parallel operations. Therefore, one approach to improving sparse compute efficiency is to enforce some degree of structure on the sparsity pattern to better align it with the hardware architecture. This has motivated a wide range of \textit{structured} pruning techniques such as  neuron~\citep{Ebrahimi2021}, channel-wise~\citep{He2017}, block~\citep{Gray2017}, 2:4~\citep{mishra2021}. These methods however almost always pay some task performance penalty compared to equivalent unstructured methods leading to a trade-off between model size and execution speed which is still not guaranteed to achieve practical performance benefits.

In this work we present PopSparse, a library for the effective acceleration of sparse matrix multiplication (matmul) on Graphcore IPUs~\citep{BowIPU}. 
IPUs have a number of architectural features that help accelerate sparse operations, firstly they have a large amount of high bandwidth on-chip SRAM which drastically improves performance for communication heavy, low arithmetic efficiency operations. This is critical for sparse operations as the increased ratio of communication to compute is a natural consequence of weight sparsification. Furthermore, they use a fine grained MIMD processing model that partitions work over 1472 independent compute units, here on described as \textit{tiles}. This means high degrees of parallelism can be exploited from operations even if they cannot be structured as a single large matrix multiplication. Finally, the IPU uses an ahead-of-time compilation model that allows further optimisations to be employed if computational details, like the structure of the sparsity, are known at compile time. PopSparse leverages these advantages to accelerate sparse matrix multiplications on IPU and allows further performance improvements to be realised by enforcing algorithmic constraints to the block size and ensuring sparsity patterns are available at compile time.

We benchmark PopSpare performance on sparse-dense matmul (SpMM) across a range of problem specifications and compare performance to dense matmul implementations on IPU as well as GPU baselines. 

\subsection{Contributions}\label{Contributions}
The present work makes the following contributions:
\begin{itemize}
\item We introduce the PopSparse library for sparse tensor operations on IPU. The library supports two flavours of sparse operations: static, where the sparsity pattern for the sparse operand is fixed; and dynamic, where sparsity pattern for the sparse operand is updated at runtime.
\item We compare benchmark results for static and dynamic sparse modes with a single sparse-dense SpMM against the dense implementation on IPU and both sparse (cuSPARSE\textsuperscript{\texttrademark}, \citet{cusparse}) and dense (cuBLAS\textsuperscript{\texttrademark}, \citet{cublas}) implementations on GPU, showing that PopSparse static sparsity outperforms competitive dense implementations in FP16 with low sparsity (ranges from 97~$\%$ to 88~$\%$, depending on the matrix size and block size). 
\item Reviewing the benchmark results, we answer the questions: 1) Under what conditions do we expect to benefit from sparse operations on IPU? and 2) How well does IPU performance compare to GPU in sparse operations?
\end{itemize}

\subsection{Related work}\label{Related-work}
Various algorithms have been proposed for SpMM acceleration on GPUs. \citet{Yang2018} introduced novel algorithms (MergeSpMM) for sparse-dense matrix multiplication on the GPU with merge-based load balancing and row-major coalesced memory access.
\citet{Hong2019} proposed an adaptive sparse tiling (ASpT) technique and used it to enhance the performance of SpMM. They used intra-row reordering to enable adaptive tiling.
~\citet{Gale2020} designed high-performance SpMM kernels
targeted specifically at deep learning applications.

Using SpMM as the fundamental component, pruning has been used to optimally sparsify the dense weight matrix in deep neural networks. \citet{Zhu2017} pruned the large models in an unstructured way and compared the accuracy of the models with their small-dense counterparts across a broad range of neural network architectures (CNN, LSTM etc.) and found that large-sparse models to consistently outperform small-dense models and achieve up to 10x reduction in number of non-zero parameters. \citet{Yao2018} showed a novel fine-grained unstructured sparsity approach and corresponding pruning method to fixing the number of non-zeros in small regions of the sparse matrix in matrix row. \citet{Wang2020} presented SparseRT, a system to support efficient inference with unstructured weight pruning on GPU. 

However, unconstrained magnitude pruning resulting in unstructured sparsity patterns is challenging to support on traditional hardware accelerators~\citep{Narang2017, Yao2018, Wang2020}. This discovery leads to structured pruning methods that results in block sparsity.
\citet{Anwar2015} introduced structured sparsity at various scales (channel wise, kernel wise and intra kernel strided sparsity) for convolutional neural networks.
Focusing on CNN area, a new channel pruning method of structured simplification to accelerate very deep convolutional neural networks is introduced by \citet{He2017}. 
\citet{Gray2017} developed highly optimized GPU kernels for neural network training and inference with block sparsity.
\citet{Lin2019} proposed structure sparsity
regularized (SSR) filter pruning scheme to speedup the computation
and reduce the memory overhead of CNNs and it was deployed to a variety of state of the art CNN structures.
Mixture-of-Experts (MoE) training computation in terms of block-sparse
operations was employed by ~\citet{Gale2022} and they also developed new block-sparse GPU kernels that efficiently handle the dynamism present in MoEs.

\section{IPU architecture}\label{IPU-architecture}
Intelligence Processing Unit (IPU) is an accelerator for machine intelligence, made up of 1,472 independent compute cores called \textit{tiles} which co-locate hardware compute logic with local SRAM.
Each tile has 625KB of local on-chip SRAM memory, leading to 900MB in total for one IPU. High-bandwidth, low-latency communication among the tiles is achieved through an all-to-all exchange fabric. 
The IPU supports a bulk synchronous parallel (BSP) execution model where each tile first does \textbf{compute} with the data available on its local SRAM followed by a stage \textbf{synchronizing} with other tiles across all IPUs, finally \textbf{exchanging} data with other tiles on the same IPU or on a different IPU~\citep{Helal2022}. The efficient communication, fine-grained independent processing and high level of parallelism are the key factors that enable fast sparse operations.

Programs run over a set of IPUs and follow a path of specified control flow. They manipulate variables just like standard programs on a CPU. The variables being manipulated are large arrays of data (tensors). The programs manipulate these variables with a set of highly parallel tasks (called vertices) executed on the threads of the tile processors. These sets of tasks are known as compute sets~\citep{IPUprogramming}.

The vertices from the multiple compute sets, and the data they manipulate in the programs form a computational graph. The computational graph, which we call Poplar~\citep{Poplar} graph, holds the relationship between compute sets and data. The computational graph gets compiled to binary code using Poplar compilation tools on the host and then gets loaded onto the IPU. The duration of converting the computational graph to binary code is called the compile time. Once the programs are loaded and the IPU configured, a host can then instruct the IPUs to run these programs. This is defined as runtime.

\section{PopSparse implementation}\label{Methodology}
PopSparse and the equivalent GPU baselines implement multiple sparse operations, however our focus will be on a sparse-dense matmul (SpMM), written:
\begin{align*}
    Y &= (M \odot W) * X \,,\\
    M_{ij} &= \hat{M}_{\lfloor i/b \rfloor, \lfloor j/b \rfloor} \,,
\end{align*}
where $\odot$ denotes elementwise multiplication, $*$ for inner product, $Y \in \mathbb{R}^{m \times n}$, $X \in \mathbb{R}^{k \times n}$ are dense output and input matrices respectively. The sparse weight matrix $(M \odot W)$ is defined via $M \in \mathbb{B}^{m \times k}$ ($\mathbb{B} = \{0, 1\}$), a mask that represents the \textit{sparsity pattern}, itself derived from $\hat{M} \in \mathbb{B}^{\lceil m/b \rceil \times \lceil k/b \rceil}$, a block mask and $W \in \mathbb{R}^{m \times k}$ defines weight values.

In this formulation, $(M \odot W)$ has a block-sparse structure, where contiguous square blocks of weights of shape $b \times b$ contain all the non-zero elements, given \textit{block-size} $b$. The dimensions $m$ and $k$ are referred to as output and input \textit{feature size} and  $n$ is referred to as \textit{batch size}, corresponding to their typical role in weight-sparse neural network computation.

We define \textit{density}, $d = \sum_{ij} M_{ij} / (m \cdot k)$, where the standard term \textit{sparsity} corresponds to $(1 - d)$. We use floating point operation (FLOP) count to quantify the amount of useful arithmetic work in an SpMM as: $2\cdot m\cdot k\cdot n\cdot d$. Note that this only counts operations performed on non-zero values and does not depend on block size $b$.

\subsection{Overview}
The PopSparse library supports unstructured (block size $b = 1$) or structured sparsity ($b = 4, 8, 16$). Only blocks up to size $b = 16$ are explicitly considered as this already achieves the theoretical maximum FLOP rate for the on-tile compute operations and we expect larger block sizes to negatively impact task performance~\citep{Dietrich2021}. Furthermore, larger blocks could be made up of many 16x16 blocks if desired. Two sparsity modes, static and dynamic, are available. Static sparsity has the sparsity pattern $M$ fixed at compile time while with dynamic sparsity only the maximum density $d^{\mathrm{max}}$ is fixed. Consequently, additional planning is required for dynamic sparsity during compilation and there is an overhead for partitioning based on a specific sparsity pattern during runtime in order to balance the quantity of sparse elements across tiles.

\subsection{Static sparsity}\label{static_sparsity}
With static sparsity, during compile time, the sparsity pattern and matrix shapes ($m$, $k$ and $n$) are known to the partitioner, which splits the non-zero elements (specific values not known yet) of the sparse matrix across $k$ dimension into $q^k$ partitions and the dense matrix across $n$ dimension into $q^n$ partitions. Where the total number of partitions, $q^k \cdot q^n$, is upper bounded by the number of tiles.
Figure~\ref{fig:algorithm_static} illustrates the distribution, computation and reduction of the sparse matrix in static sparse. Splits over the $k$ dimension do not have to be evenly sized, and are chosen to ensure a balanced distribution of the non-zero elements (denoted with coloured small blocks in the figure). In the diagram, three distributed partitions of non-zero elements are mapped onto three tiles. A Poplar~\citep{Poplar} graph (compute graph) is then constructed based on the partition information. 

At runtime, the specific non-zero values of $W$ are provided by the host and re-ordered to match the distribution of non-zero elements on tiles of the IPU. As this distribution is known in advance, no extra exchange of data is needed. Local dot product computation with the dense matrix is then executed and there is a final reduction across tiles to give the output dense matrix. A flow chart for static sparsity can be found in Figure~\ref{fig:static_flow} of the Appendix~\ref{AppendixA-1}.

\subsection{Dynamic sparsity}\label{dynamic_sparsity}
With dynamic sparsity, the maximum number of non-zero elements for the sparse operand is fixed, while the sparsity pattern is updated during runtime. To minimise compute cycles, we employ a planner during compile time to optimise the partitioning while accommodating for the full range of sparsity patterns that can be defined by the host at runtime. Specifically, the planner determines how many \textit{equal} parts each dimension ($m, k, n$) should be divided into with each tile being assigned to a single partition. 

At runtime, as the sparsity pattern could change for each run, compared to static sparsity, there is an additional distribution of sparse matrix indices and non-zero values by the partitioner. This is demonstrated in Figure~\ref{fig:algorithm_dynamic}. Furthermore, as fixed sized partition over $m$ and $k$ dimension is required, the number of non-zero elements in each partition may not be balanced. In Figure~\ref{fig:algorithm_dynamic} the B elements on the second partition of $k$ dimension do not fit on $\mathrm{Tile 1}$. Three of the B elements overflow to $\mathrm{Tile 2}$. However, the $\mathrm{Tile 2}$ does not have all the partition information (slices of $m, k, n$) corresponding with B elements to compute the result. Consequentially, extra steps of propagation are induced to finish the computation. A dynamic number of steps is required (depending on the sparsity pattern) to propagate information and compute the final result.

Figure~\ref{fig:dynamic_flow} in Appendix~\ref{AppendixA-1} shows a diagram of the dynamic sparse-dense matmul operation during compile time and runtime. More details about dynamic sparsity can be found in Appendix~\ref{AppendixA-2}.

\begin{figure}[t]
    \centering
    \subfloat[Static]{\label{fig:algorithm_static}
        \centering
        \includegraphics[width=0.5\textwidth]{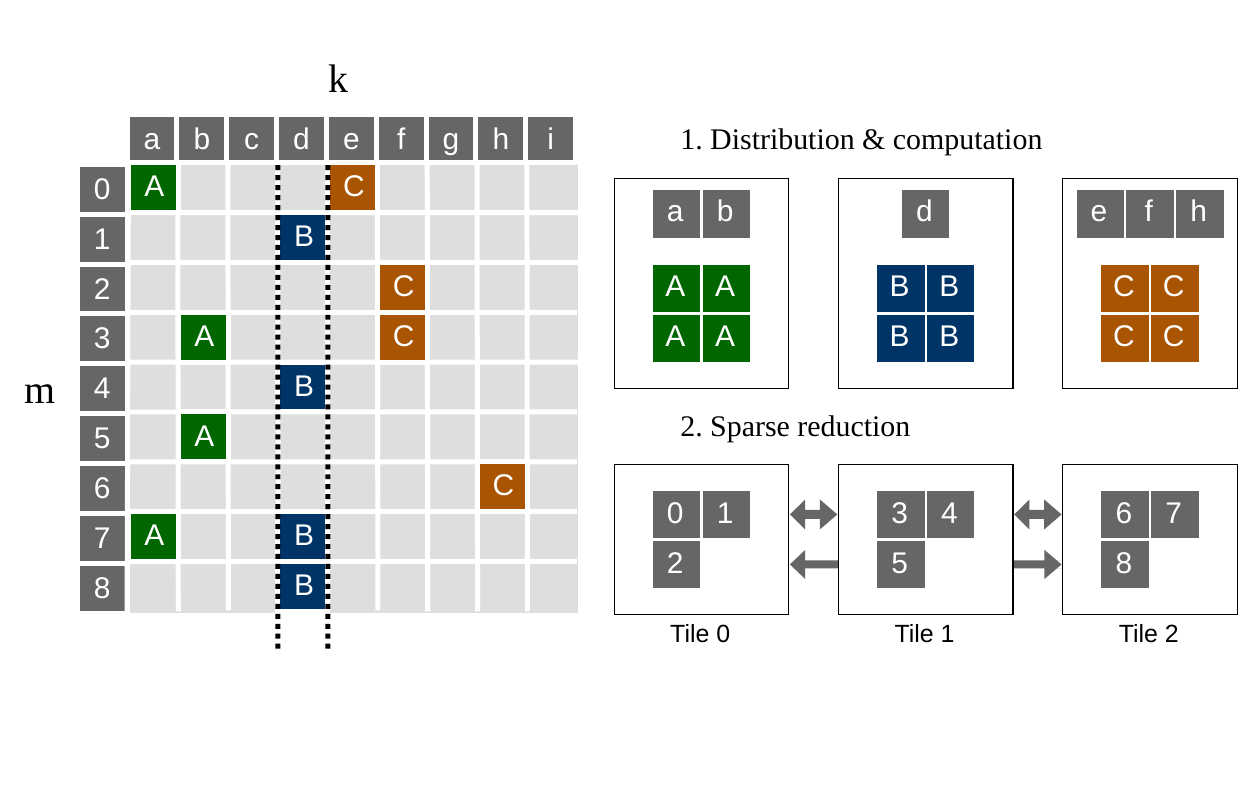}
    }
    \subfloat[Dynamic]{\label{fig:algorithm_dynamic}
        \centering
        \includegraphics[width=0.5\textwidth]{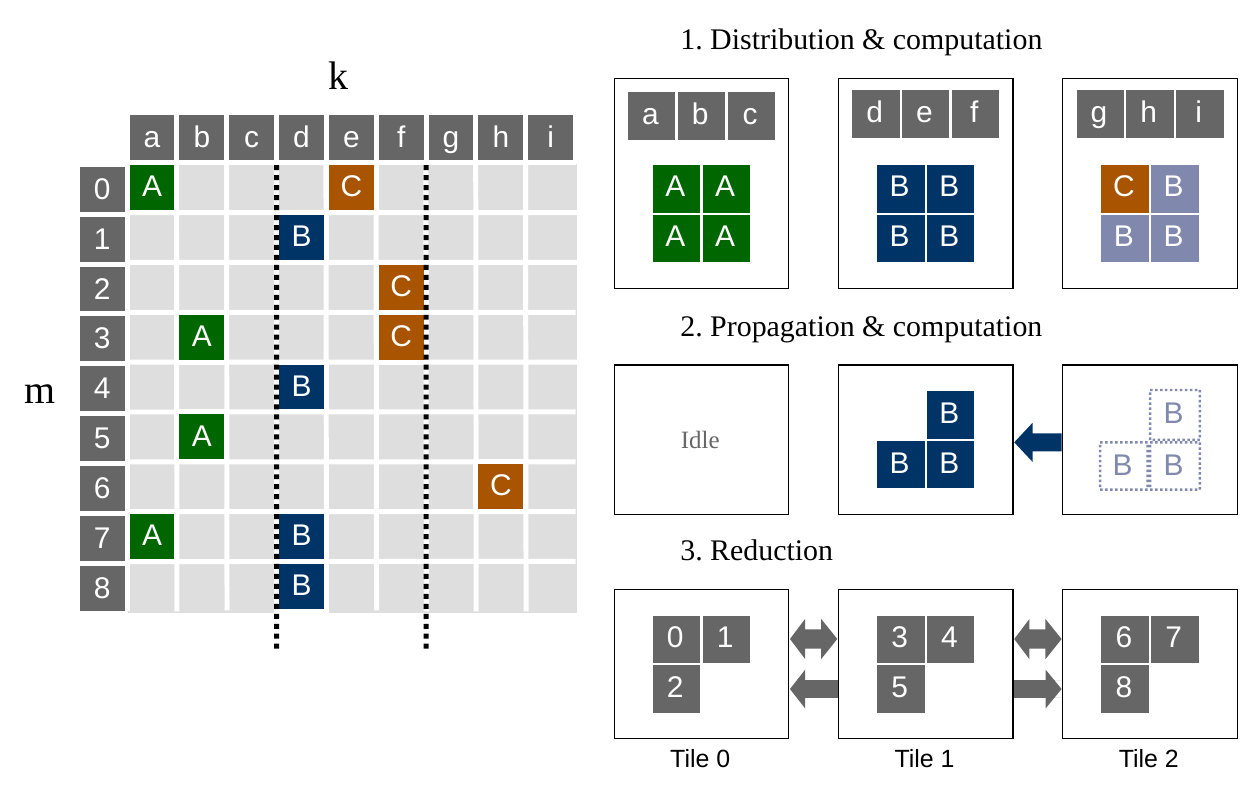}
    }
    \caption{Illustration of static and dynamic sparse partitioning of the input feature dimension, $k$. Note that $m\!=\!k\!=\!9$ and we split work over 3 tiles with $q^k\!=\!3$, $q^m\!=\!q^n\!=\!1$. Static partitioning can adapt the split positions to the known sparsity pattern, removing the need for dynamic sparsity's allocation overflow (b.1) and redistribution phase (b.2). Static execution also allows optimal exchange of inputs, (a.1) vs (b.1) and in output reduction, (a.2) vs (b.3).}
\end{figure}

Overall our dynamic sparse implementation has a number of additional costs compared to the static case:
\begin{itemize}
\item On-tile compute requires additional control flow which incurs some cost overhead.
\item Exchange phases must account for the largest communication volume possible and thus are less efficient.
\item When data is unbalanced, additional propagation and compute phases are required to rebalance the data. The number of steps required is dependent on the degree of the imbalance.
\end{itemize}

\begin{table}[t]
    \caption{APIs used for benchmarking.}
    \label{tab:apis}
    \vspace{-5pt}
    \centering
    \fontsize{8.5pt}{8.5pt}\selectfont
    \renewcommand{\arraystretch}{1.2}
    \newcommand{\tcode}[1]{\fontsize{7pt}{9pt}{\texttt{#1}}}
    \begin{tabular}{llllll}
    \toprule
    \textbf{Device} & \textbf{Implementation} & \textbf{API} & \textbf{Block size} & \textbf{Dynamic} & \textbf{Datatype support} \\
    \midrule
    \rowcolor{lightgray!20}IPU & Dense & \tcode{poplin::matMul} & -- & -- & FP16, FP32 \\
    IPU & Static sparse (\tcode{popsparse::}) & \tcode{static\_::sparseDenseMatMul} & 1,4,8,16 & \xmark & FP16, FP32 \\
    \rowcolor{lightgray!20}IPU & Dynamic sparse (\tcode{popsparse::}) & \tcode{dynamic::sparseDenseMatMul} & 1,4,8,16 & \cmark & FP16, FP32 \\
    GPU & Dense & \tcode{cublasGemmEx} & -- & -- & FP16, FP32 \\
    \rowcolor{lightgray!20}GPU & Compressed Sparse Row (CSR) & \tcode{cusparseSpMM} & 1 & \cmark & FP16$^{*}$, FP32 \\
    GPU & Block Sparse Row (BSR) & \tcode{cusparseSbsrmm} & $2^{b'}$ & \cmark & FP32 \\
    \bottomrule
    \multicolumn{3}{l}{\rule{0pt}{1em}\footnotesize{$*$ compute in FP32, inputs/outputs in FP16}}
    \end{tabular}
\end{table}

\section{Benchmarks}\label{Benchmarks}
Our microbenchmarking experiments explore a single sparse-dense matmul SpMM: $Y = (M \odot W) * X$, on IPU and GPU. The APIs used for IPU experiments and GPU baselines are shown in Table~\ref{tab:apis}. The range of parameters that were benchmarked is detailed in Table~\ref{parameters-range}.

\begin{table}[!h]
    \caption{Ranges of parameters swept for benchmarks.}
    \label{parameters-range}
    \vspace{-5pt}
    \centering
    \fontsize{8.5pt}{8.5pt}\selectfont
    \renewcommand{\arraystretch}{1.2}
    \begin{tabular}{llll}
    \toprule
    \rowcolor{white}\textbf{Parameter}           & \textbf{Range (start:step:end or list of values)}  \\ \midrule
    \rowcolor{lightgray!20}feature size ($m=k$) & $2^{8:1:13}$\\
    batch size ($n$)  &$2^{2:2:16}$  \\
    \rowcolor{lightgray!20} block size ($b$)    &$1$ (unstructured), $4, 8, 16$   \\
    density factor ($d$)  &$1$ (dense), $1/4, 1/8, 1/16, 1/32$  \\
    \rowcolor{lightgray!20} data type             &FP16, FP16$^*$, FP32 \\
    \bottomrule
    \multicolumn{2}{l}{\rule{0pt}{1em}\footnotesize{$*$ compute in FP32, inputs/outputs in FP16}}
    \end{tabular}
    \vspace{-15pt}
\end{table}

\paragraph{IPU}
The operation is executed a single time on a single Bow IPU~\citep{BowIPU} in a Bow-2000 chassis, with randomly generated sparsity pattern and values. We extract cycle count information and convert these cycle counts into TFLOP$/$s values given a constant clock of 1.85~GHz \citep{BowSpec}. Host transfers are excluded from measured time.

\paragraph{GPU baselines}
Sparse implementations on GPU are provided by the cuSPARSE library\footnote{cuSPARSE version 11.6.2}~\citep{cusparse}. These are divided according to the sparse format used, with no distinction between static and dynamic sparsity patterns. We consider compressed sparse row (CSR) and block sparse row (BSR), for unstructured and block sparsity respectively.\footnote{We skip the detailed implementation here for brevity, referring to \citet{Yang2018} and \citet{Huang2020} for a discussion of such considerations (although these do not discuss the implementation of cuSPARSE directly).} There are other vendor-provided and third-party libraries for sparse computation, but we consider these for sake of direct comparison against PopSparse (see Appendix~\ref{AppendixB} for more information). Dense benchmarks used cuBLAS~\citep{cublas}.

To measure runtime, we generate a random sparsity pattern and values, copy to device, execute 25 iterations of the operation and copy results to host for validation. We start timing after 5 iterations, and measure wall clock time using \texttt{cudaEventRecord}, at millisecond resolution, over all remaining iterations. We use \texttt{cudaDeviceSynchronize} before the start and after the stop of wall clock recording to ensure previous matmul operation is complete. All experiments were performed on an A100 GPU running in a 4 x A100-SXM4-40G chassis with CUDA version 11.3.

\section{Results}

In our benchmarking experiments, we explore the performance of static and dynamic sparsity on IPU, compared with dense computation on IPU and both dense and sparse computation on GPU.

For most results, we allow batch size $n$ to vary, choosing the batch size leading to best performance for each configuration. This is appropriate for batched inference in a weight sparse neural network, as batch size is a free parameter (up to a memory limit) and the choice depends on the machine learning application. Note that we consider only non-zero elements in the sparse matrix to calculate FLOP$/$s, as defined in Section~\ref{Methodology}.

\subsection{Dense baseline performance}\label{baseline}
Figure~\ref{fig:dense-ipu-gpu} shows dense matmul performance, as batch size and feature size vary for FP16 and FP32. In FP16, we see chip-for-chip parity between IPU and GPU at large batch size, while the IPU is more resilient to low batch size. This is because lower arithmetic intensity operations leads to more data movement with the same amount of compute. The SRAM on IPU enables more efficient data movement.

In FP32, the IPU has a clear advantage due to Accumulating Matrix Product (AMP) units~\citep{AMP} being available in FP32, whereas Tensor Cores~\citep{tensorcores} are only available in FP16.

\begin{figure}[t]
    \centering
    {\includegraphics[width=\textwidth]{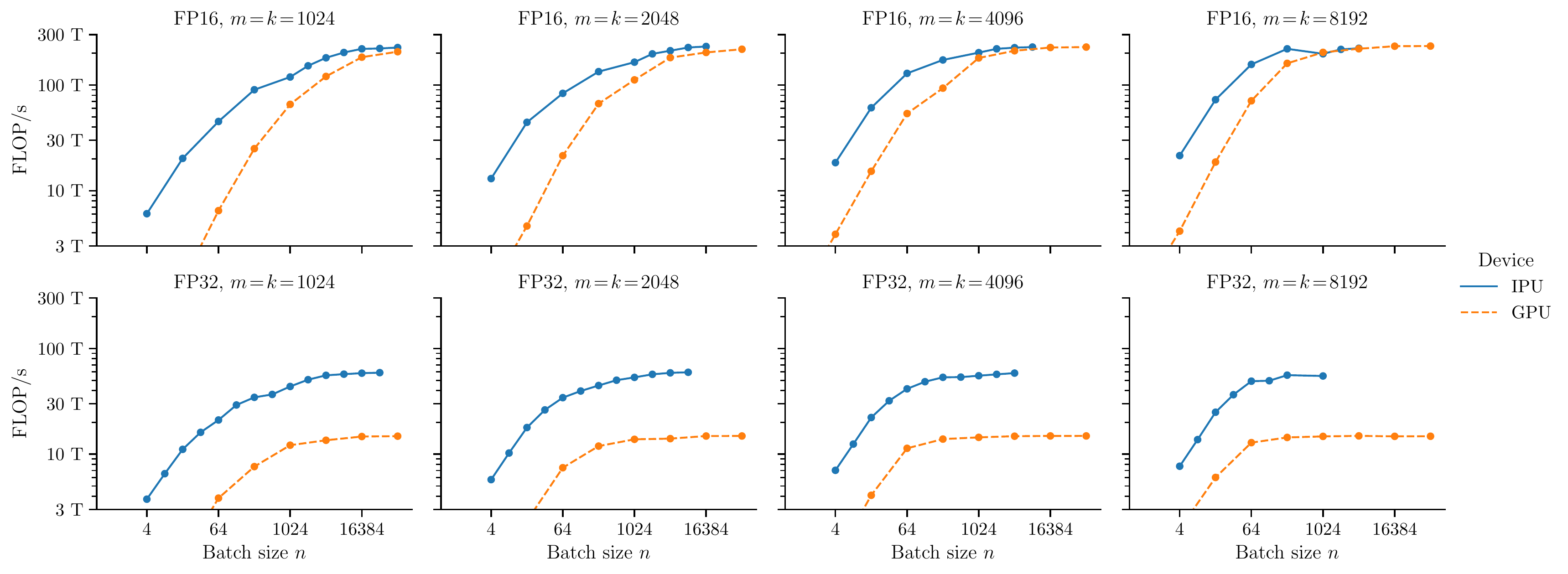}}
    \caption{Dense performance for large square matrices on IPU and GPU.}
    \label{fig:dense-ipu-gpu}
\end{figure}

\subsection{Static vs. dynamic sparsity on IPU}\label{static-dynamic-ipu}
It has already been discussed in Section~\ref{dynamic_sparsity} that static sparse performance should exceed dynamic. This is visible in Table~\ref{static_vs_dynamic}, showing the speedup of dynamic and static sparsity versus dense for a variety of configurations. Over various block sizes and data types, static sparsity shows higher speedup than dynamic. The sparse computation speedup also increases with block size. More discussion can be found in Section~\ref{sparse-dense-ipu}. Moreover, we can see that the sparsity speedup for FP32 is better than for FP16. This is because core arithmetic is more expensive in FP32, so FLOP savings count more (versus sparse overhead).

\begin{table}[!h]
\caption{Dynamic sparse vs. static sparse on IPU, $m = k = 4096$, density $d = 1/16$, best over batch size $n$, throughput values compared with dense.}
\vspace{-10pt}
\label{static_vs_dynamic}
\vspace{5pt}
    \centering
    \fontsize{8.5pt}{8.5pt}\selectfont
    \renewcommand{\arraystretch}{1.2}
    \setlength\tabcolsep{1pt} 
    \begin{tabular}{llll}
    \toprule
    \rowcolor{white}\textbf{Block size} \hspace{10pt} & \textbf{Type} \hspace{10pt} & \textbf{Dynamic/dense}  \hspace{10pt} & \textbf{Static/dense}  \\ \midrule
    \rowcolor{lightgray!20}1         &FP16    &0.4    &0.7 \\
    1             &FP32    &0.9    &1.4 \\
    \rowcolor{lightgray!20}4            &FP16     &1.0    &1.5 \\
    \rowcolor{lightgray!20}4            &FP32     &2.7    &3.2 \\
    16            &FP16     &1.9    &4.9 \\
    \rowcolor{lightgray!20}16            &FP32     &3.8    &5.6 \\
    \bottomrule
    \end{tabular}
\end{table}

\subsection{When does a sparse implementation outperform dense on IPU?}\label{sparse-dense-ipu}
Our results show that sparse implementations are faster than dense on IPU at \textit{low density}, with \textit{large block size} and \textit{large feature size}.

\begin{figure}[t]
    \centering
    \subfloat[IPU]{\label{fig:ipu-density}
        \centering
        \includegraphics[width=0.49\textwidth]{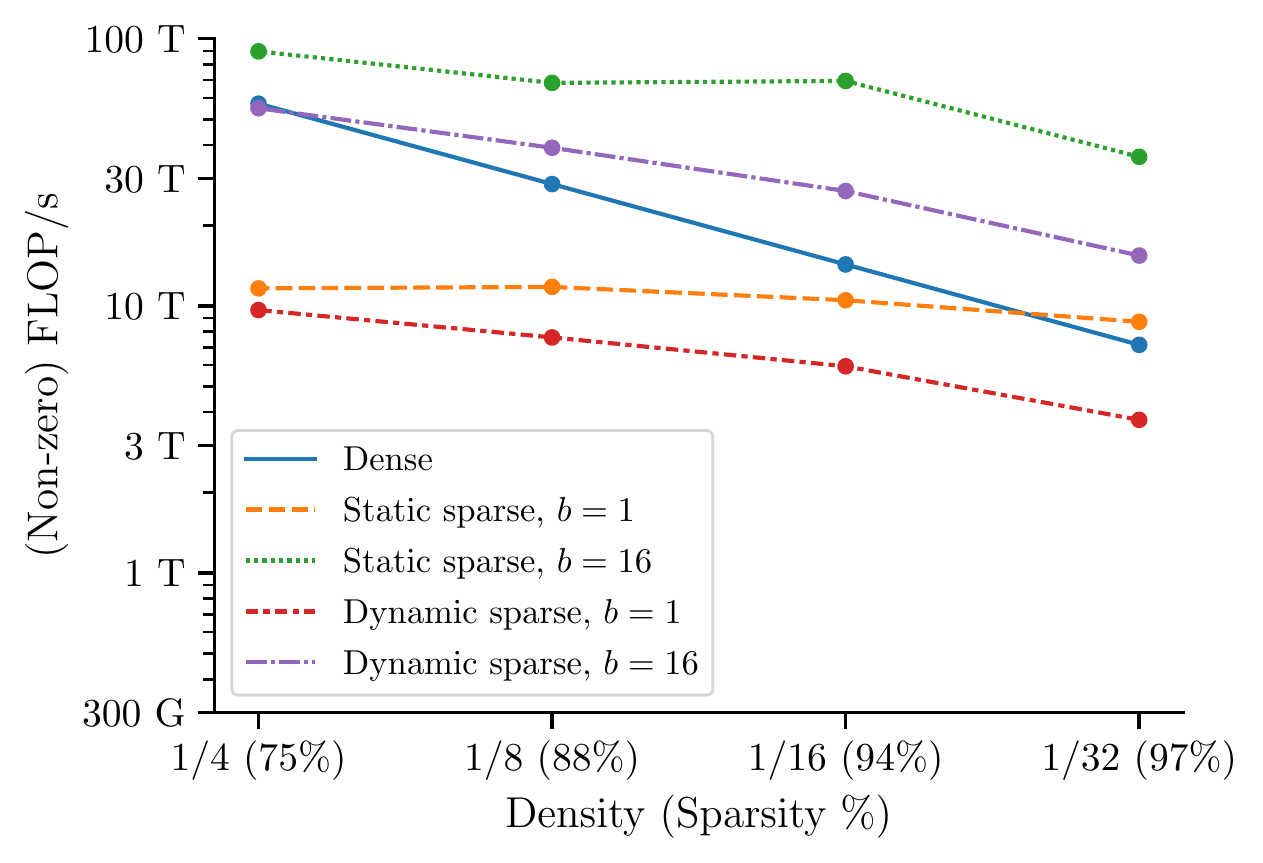}
    }
    \subfloat[GPU]{\label{fig:gpu-density}
        \centering
        \includegraphics[width=0.49\textwidth]{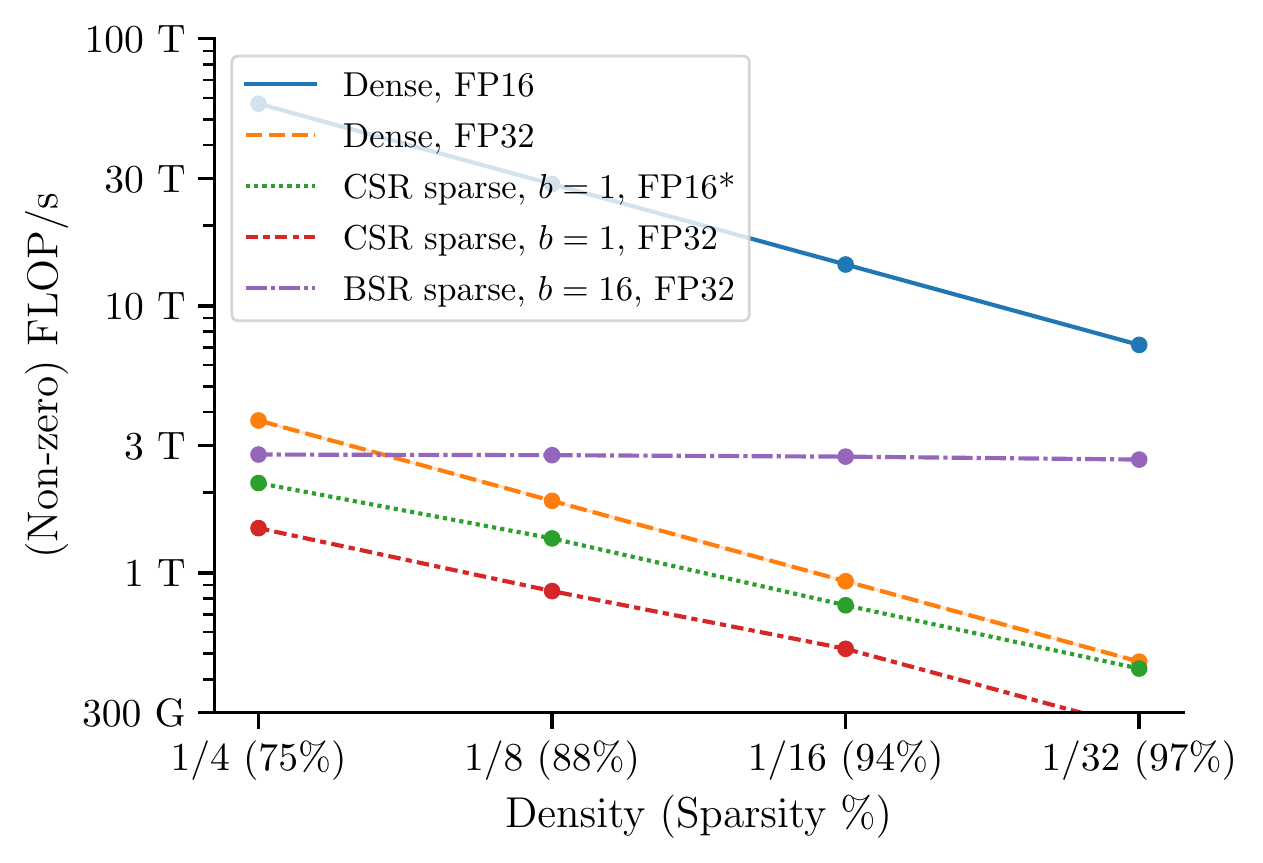}
    }
    \caption{\protect\subref{fig:ipu-density} IPU FP16 and \protect\subref{fig:gpu-density} GPU block-sparse matmul performance as density varies for various block size $b$, square feature size $m=k=4096$, best over batch size $n$.}
\end{figure}

\paragraph{Low density}
Since the purpose of sparse implementations is to take advantage of zeros, it is unsurprising that decreasing non-zero density leads to a performance advantage over dense. In Figure~\ref{fig:ipu-density}, we show the plot of compute FLOP/s (in log scale) vs. density with fixed feature size. This shows the density scaling performance of dense, dynamic sparse and static sparse for block size $b=1,16$. The dense implementation shows linear scaling with density --- since zeros are not counted in the FLOP/s calculation, the amount of useful work decreases with $d$, while the runtime is unchanged. While on the other hand for a sparse implementation, perfect scaling would predict constant FLOP/s as density varies, with both work and runtime scaling with $d$. We see nearly-perfect scaling from static sparsity in this range, whereas dynamic sparsity shows higher overheads. In Figure~\ref{fig:ipu-density}, unstructured ($b=1$) static sparsity outperforms dense at a density of 5$\%$.

\paragraph{Large block size}
Block sparsity speeds up sparse computation for two main reasons. First, there is less overhead incurred to store and process the metadata that encode the sparsity pattern. Second, it allows implementations to process non-zero submatrices rather than just vectors, making more effective use of the hardware. Figure~\ref{fig:ipu_block_size} shows the magnitude of these improvements, which are considerable: $2.1\times$ for blocks size $b=4$ and $6.6\times$ for $b=16$ (static sparsity). 
Despite this advantage, few sparsity techniques in the literature consider block sparsity, so these results could motivate the field to explore new ML algorithms for inducing blocks of zeros.

\paragraph{Large feature size}
Figure~\ref{fig:ipu_feature_size} shows the effect of feature size on sparse performance. Sparse performance improves more with feature size than dense performance. This is because increasing feature size makes it easier to spread work evenly across tiles.

\paragraph{Will my application speed up?}
The multidimensional nature of our sweep makes it hard to summarise results on a single plot. To make it easier to evaluate whether a particular problem can be accelerated using PopSparse, we provide a grid of static sparse speedup ratios in Figure~\ref{fig:ipu-static-grid}. We also fit a power-law model to our static sparse results, illustrated in Figure~\ref{fig:ipu_model}. This predicts the condition for achieving a speedup: $0.0013 \cdot m^{0.59} \cdot d^{-0.54} \cdot b^{0.50} > 1$. This is in line with our main observations: large feature size, low density and large block size all favour sparse implementations. We recommend using this for interpolation only, as we do not expect good generalisation beyond the sweep parameters of Table~\ref{parameters-range}.

\begin{figure}[t]
    \centering
    \subfloat[]{\label{fig:ipu_block_size}
        \centering
        \includegraphics[height=4.3cm]{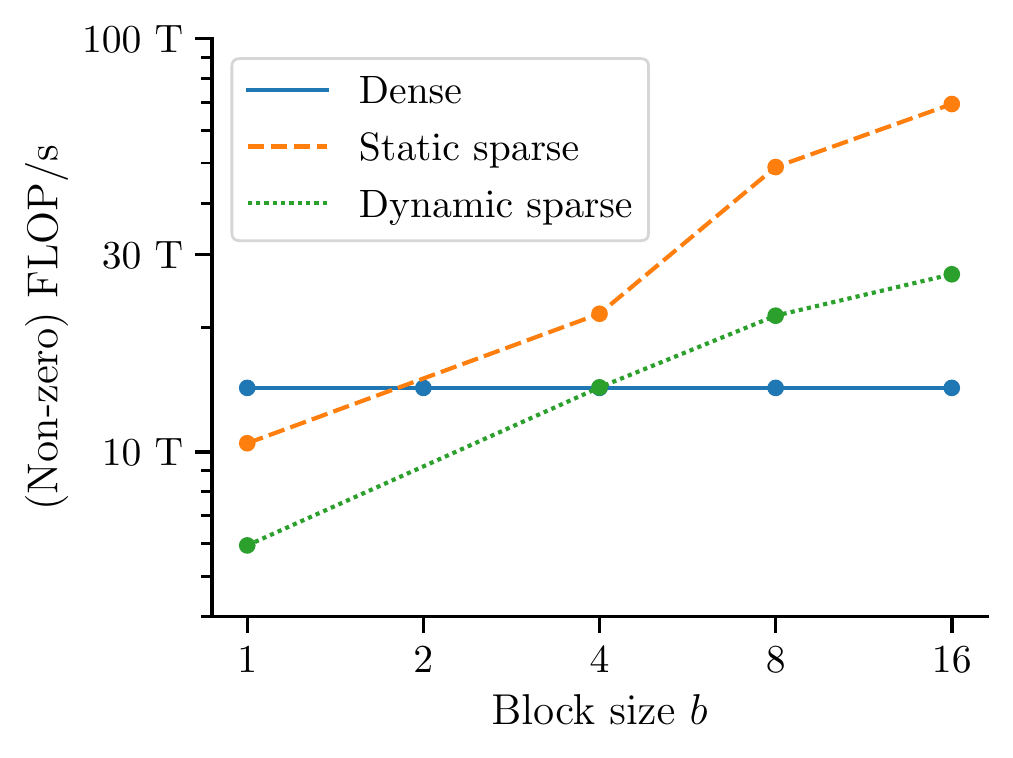}
    }
    \subfloat[]{\label{fig:ipu_feature_size}
        \centering
        \includegraphics[height=4.3cm]{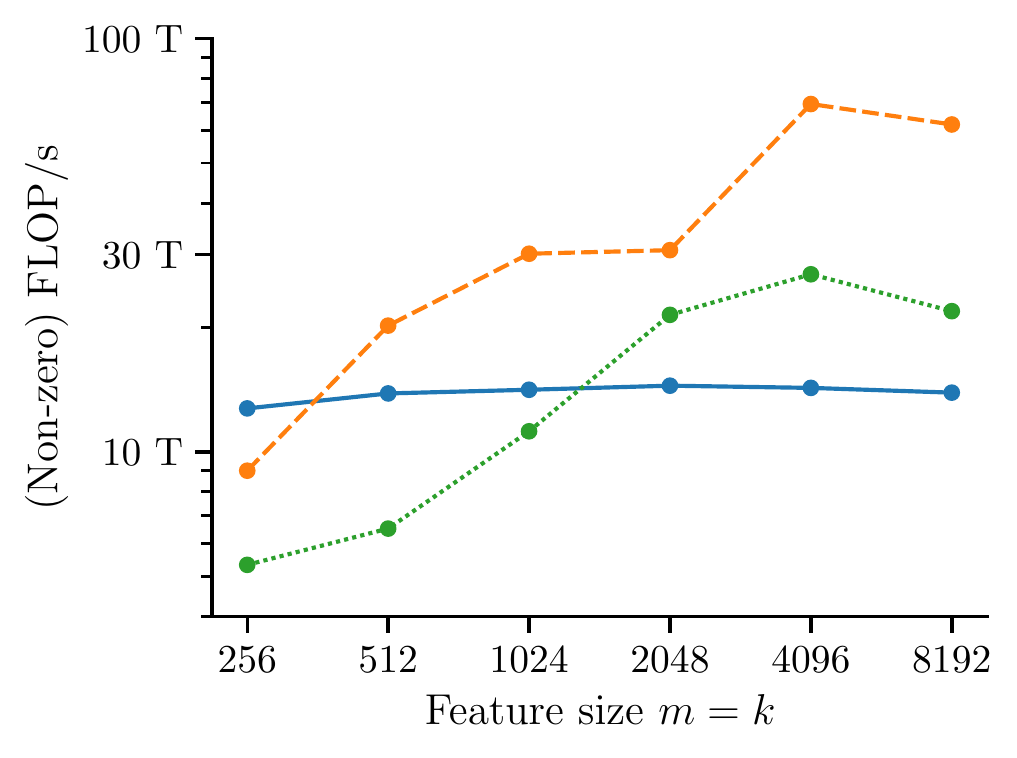}
    }
    \subfloat[]{\label{fig:ipu_model}
        \centering
        \includegraphics[height=4.3cm]{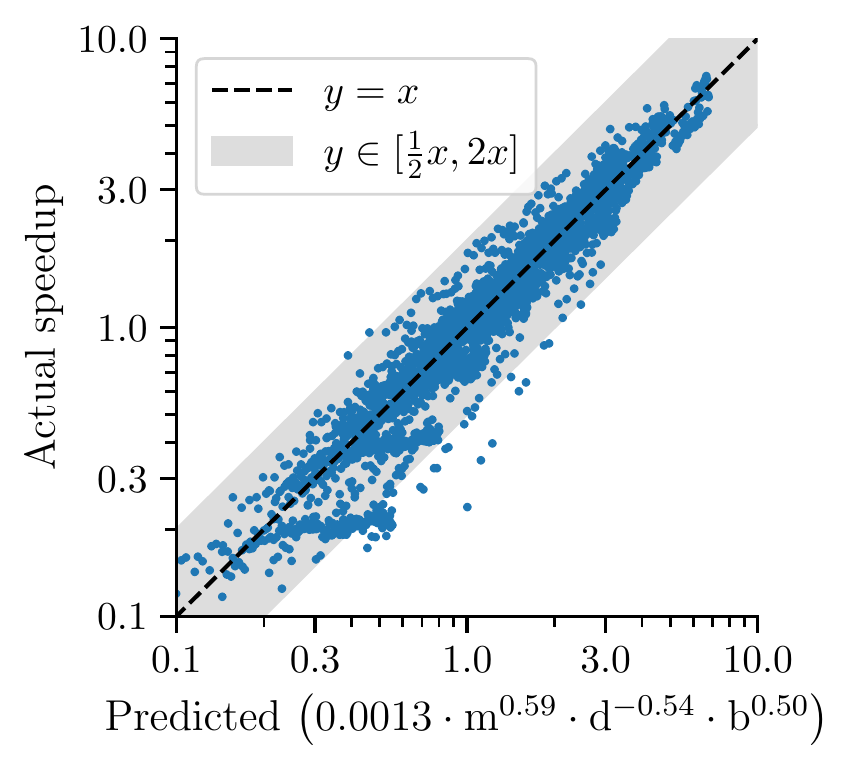}
    }
    \caption{IPU block-sparse matmul performance as \protect\subref{fig:ipu_block_size} block size or \protect\subref{fig:ipu_feature_size} feature size varies. Base configuration: FP16, $m = k = 4096$, density $1/16$, block size $16$, best over $n$.
    Figure~\protect\subref{fig:ipu_model} shows the fit of a power-law model for the sparse speedup ratio (static sparse to dense TFLOP/s), for FP16, best over $n$, for each $(m, d, b)$ tested.}
\end{figure}

\subsection{How well does IPU perform compared to GPU in sparse operations?}\label{ipu-gpu-comapre}
Figure~\ref{fig:gpu-density} shows GPU performance as density varies, showing that BSR sparsity in FP32 is worse than the FP16 dense baseline, even below $2\%$ density. However, we also observe that GPU sparse performance scales well as density decreases. We suspect the main reason for the sparse-dense performance gap is that the BSR implementation does not support FP16, and therefore cannot use Tensor Cores to accelerate the arithmetic (see Figure~\ref{fig:dense-ipu-gpu} for the FP32 versus FP16 dense performance). Note that FP16 sparsity is available in the blocked ELL format, which we did not benchmark as it did not match the API of general block sparsity and so would require explicit padding. However, it is probable that the gains due to Tensor Core acceleration would negate any padding overheads in this case, so we consider this for future work.

Since dense methods perform best on GPU, and GPU dense performs similarly to IPU dense (see Figure~\ref{fig:dense-ipu-gpu}), our previous comparisons already give an approximate answer to the question of when sparse operations on IPU outperform GPU: whenever sparse operations outperform dense operations on IPU. Therefore Figure~\ref{fig:ipu-static-grid} in Appendix~\ref{AppendixC} can also be used as an approximate guide to compare IPU sparsity with GPU.

\section{Conclusions}
In this work, we introduce the PopSparse library for sparse matrix multiplication on IPU and show the benchmarks of SpMM on IPU with static sparsity and dynamic sparsity. We also compare the sparse and dense results to a GPU baseline.

We have found that there are configurations where sparse implementations on IPU can outperform dense on IPU and all tested implementations on GPU (cuSPARSE CSR and BSR APIs).

As an approximate guide, sparsity brings speedup over dense (FP16, large batch size for both) when: 
\begin{itemize}
\item Static, block size $b=1$: features $(m=k) \ge 4096$, density $d \le 1/32$.

\item Static, block size $b\ge 4$: features $(m=k) \ge 4096$, density $d \le 1/8$.

\item Dynamic sparsity: block size $b \ge 8$, features $(m=k) \ge 4096$, density $d \le 1/32$.
\end{itemize}
The requirements for both static and dynamic sparsity to outperform dense on IPU are much more modest in FP32, however FP16 is generally preferred for weight-sparse deep learning models. 

The ranges which show good performance are somewhat challenging to use today in the context of weight sparse training and inference. Only a few deep learning literature (for example the Pixelated Butterfly from \citet{Dao2021} and Towards Structured Dynamic Sparse Pre-Training of BERT from \citet{Dietrich2021}) explore effective schemes for block sparsity. This might because finding block sparse patterns that do not degrade performance is a hard problem, and that on theoretical efficiency metrics (per non-zero FLOP), unstructured sparsity will always appear as good or better than block sparsity. We hope that this work shows that block sparsity is a promising method for achieving practical acceleration of sparse models and will motivate further investigation into effective block sparse pruning algorithms.

For the next step, we aim to demonstrate the static block sparsity to accelerate large model inference on IPU, achieving both high memory and high compute efficiency.


\bibliography{iclr2023_conference}
\bibliographystyle{iclr2023_conference}

\appendix
\section{More details about static and dynamic sparsity}\label{AppendixA}

\subsection{Flow chart for static and dynamic sparsity}\label{AppendixA-1}

\begin{figure}[!h]
    \centering
    \subfloat[Static]{\label{fig:static_flow}
        \centering
        \includegraphics[width=0.51\textwidth]{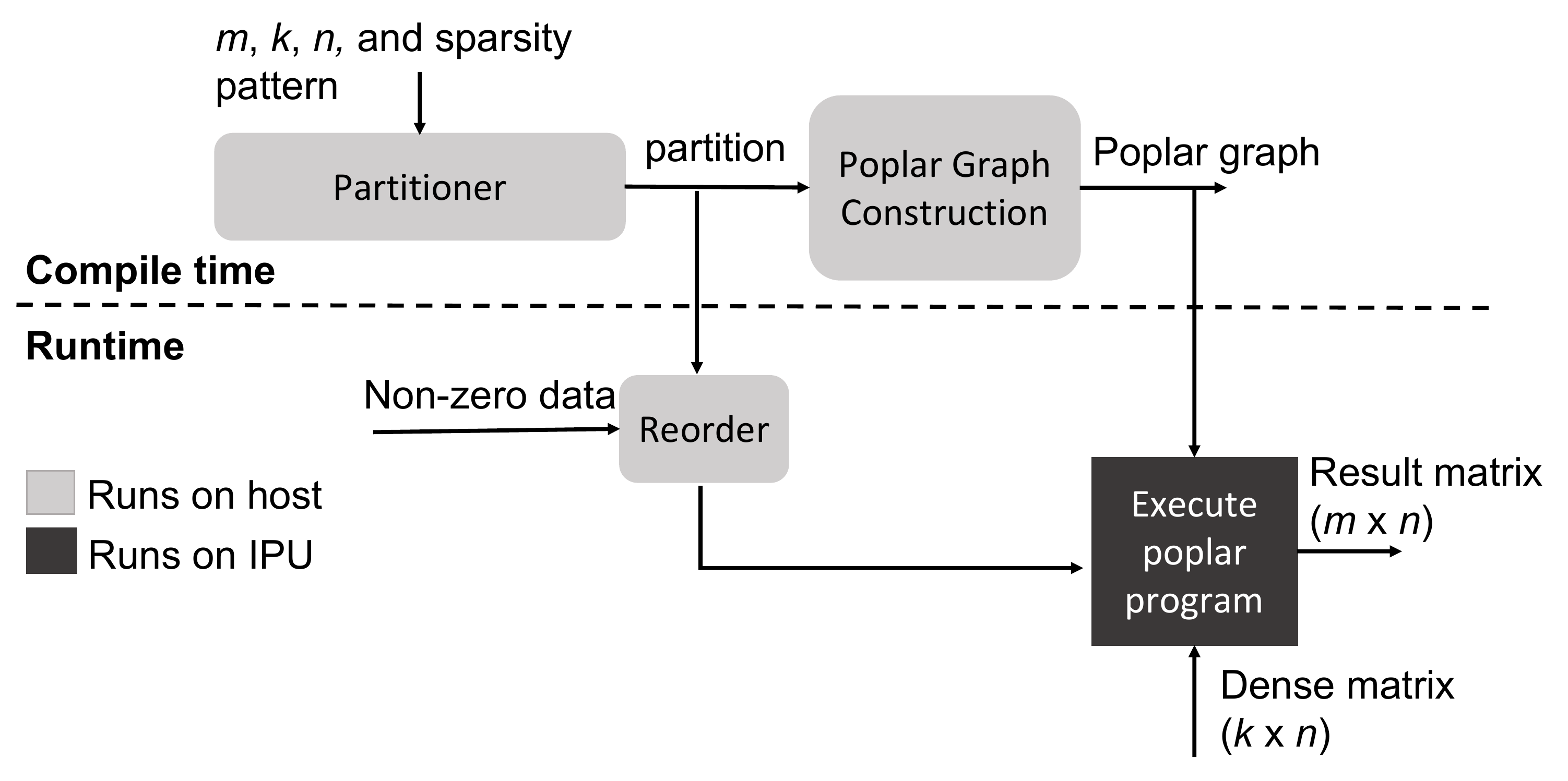}
    }
    \subfloat[Dynamic]{\label{fig:dynamic_flow}
        \centering
        \includegraphics[width=0.5\textwidth]{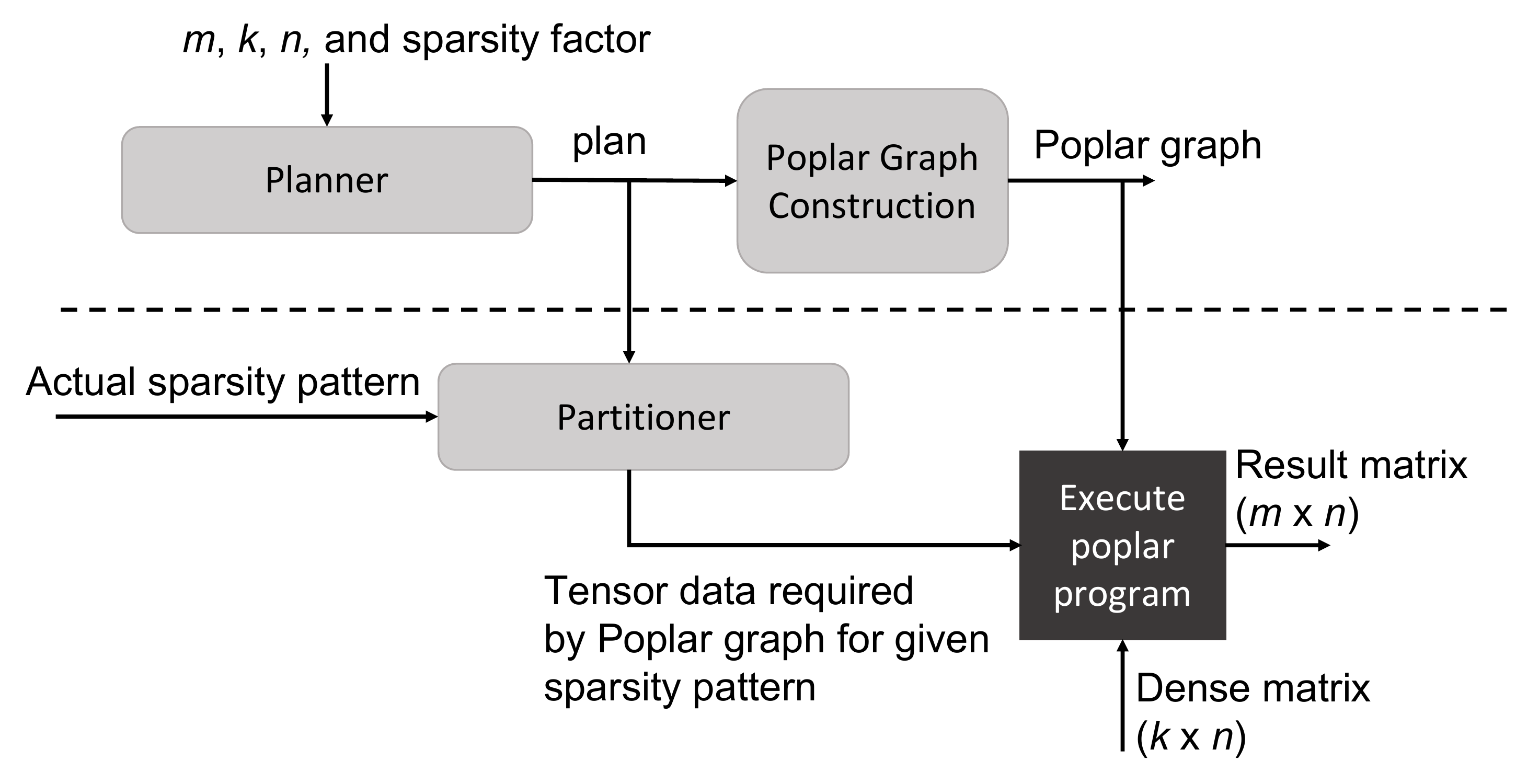}
    }
    \caption{Flow chart for static and dynamic sparsity.}
\end{figure}

\subsection{Details about dynamic sparsity}\label{AppendixA-2}
With dynamic sparsity, the total number of non-zero elements for the sparse operand is fixed (or has a known maximum) while the sparsity pattern is updated during runtime. 

There are three key components for the implementation of the dynamic sparsity in PopSparse:
\begin{itemize}
\item A \textbf{planner} which produces the plan for optimum partition and device operation.
\item A \textbf{host utility} which encodes a sparsity pattern to be uploaded to the device and used by the device-side implementation.
\item The \textbf{implementation on the device} that is based on the plan produced by the planner. This constructs the variables, compute sets and programs needed to run the ops on the device.
\end{itemize}

\textbf{The planner} consists primarily of a partition between tiles and vertices of the computation graph used on each tile. The partitioner determines how each dimension ($m, k, n$) should be divided into when calculating the result of the sparse-dense matmul. It is decided at compile time based on the matrix sizes and density factor and does not change with sparsity pattern. All partitions of each dimension are of equal size except the last which may be smaller if the number of partitions in that dimension does not evenly divide. For example, if $q^m$ is the number of partitions to divide the dimension $m$ into, there will be $q^m - 1$ partitions of size $\lfloor m/q^m \rfloor$ and the last partition with size $m - (q^m - 1) \cdot \lfloor m/q^m \rfloor$. The total number of partitions, $q^\mathrm{total} = q^m \cdot q^k \cdot q^n$.

Each tile will be assigned a single partition, for which it is responsible for computing its portion of the partial result. Therefore the number of tiles used is equal to $q^\mathrm{total}$.

\textbf{The host utility} encodes the sparse matrix row and column indices as $\mathrm{metaInfo}$. The $\mathrm{metaInfo}$ along with the non-zero values of the sparse matrix ($\mathrm{nzValues}$) are splits into buckets. The size of these buckets is set by the planner based on the maximum density that was fixed at compile time. The buckets are mapped to each of the $q^\mathrm{total}$ number of partitions so that all partitions that are assigned to tiles have equal and fixed size buckets with $\mathrm{metaInfo}$ and $\mathrm{nzValues}$. The bucket size is based on distributing the total number of non-zero values for the sparse tensor across $q^m \cdot q^k$ partitions evenly: 
\begin{equation*}
    N_{\mathrm{non\mbox{-}zero}} = m \cdot k \cdot d^{\mathrm{max}}/(q^m \cdot q^k) \,,
\end{equation*}
where $N_{\mathrm{non\mbox{-}zero}}$ is the average number of non-zero elements in a single bucket on a single tile, based on maximum density $d^{\mathrm{max}}$. The buckets are repeated over $q^n$ partitions. A similar estimate is done for the size of $\mathrm{metaInfo}$ for a single bucket but since we have a varied sparsity pattern, some extra headroom is given in the size of these buckets.

\textbf{With the device implementation}, we have compute graph vertices created on each tile with:
\begin{itemize}
\item The slice of the dense input required to compute this partition's result.
\item The bucket of the input sparse tensor mapped to this tile (both $\mathrm{metaInfo}$ and $\mathrm{nzValues}$).
\item The slice of the dense output partials that will hold the results for this partition.
\end{itemize}
The execution steps can be summarized as following:
\begin{itemize}
\item Distribution phase: The tile reads through the $\mathrm{metaInfo}$ and $\mathrm{nzValues}$ in the bucket on a tile that fall within the partition to be processed on this tile. If all the non-zero values for all $q^m \cdot q^k$ partitions are distributed in buckets on tiles assigned the same partition, the calculation is complete. If not, a further series of dynamic exchange + compute steps (propagation) are executed to complete the calculation.
\item Propagation phase, use exchanges to move buckets between tiles. This phase repeats until compute is complete.
    \subitem A, Check information encoded in $\mathrm{metaInfo}$ to see if the calculation is complete, exit if complete.
    \subitem B, If not, exchange buckets to shift between tiles.
    \subitem C, Compute, dense input is the same but the contents of the buckets on tiles have changed.
\item Reduce partial results across $q^k$ to get the final output.
\end{itemize}
The distribution and propagation phases are done with the dynamically executed steps where the number of steps are decided by the partitioner for a given sparsity pattern.

To further illustrate the distribution and propagation phases, we use two extreme cases. The best case scenario is when non-zero values in the sparse operand are evenly spread among $q^m \cdot q^k$ partitions. In this balanced case, the buckets mapped to tiles assigned each $q^m \cdot q^k$ partition should have space enough to hold all non-zero values falling within that partition. Then we can complete the operation in the distribution phase and no further propagation phases are needed. At the same time, each tile does an equal share of the total FLOPs. The worst case scenario is when all non-zero values in the sparse operand lie within a single $q^m \cdot q^k$ partition. The first tile in which all the non-zero values lie must do all the FLOPs for the operation, and all the other tiles do no useful work. Moreover, the bucket size could only take $1/(q^m \cdot q^k)$ of the non-zero values. As a result, all the non-zero elements must be spread between all the buckets. This means we must exchange + compute up to $q^m \cdot q^k$ times to complete the operation. 

To give an example, in Figure~\ref{best_scenario}, the best scenario, there are two non-zero values for each partition (in total four partitions). Each bucket contains two $\mathrm{nzValues}$ and no further exchange is needed. With the worst scenario, as denoted in Figure~\ref{worst_scenario}, all the non-zero values lie within the first tile and all the rest three tiles do no useful work.
\begin{figure}[!h]
    \centering
    \subfloat[][\centering Balanced distribution]{{\includegraphics[width=3cm]{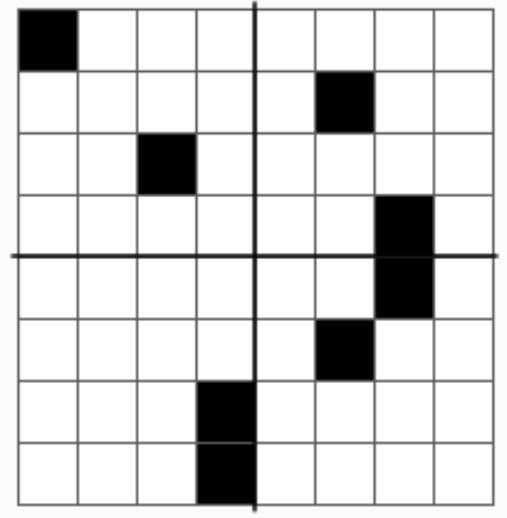}}\label{best_scenario}}
    \qquad
    \subfloat[][\centering Unbalanced distribution]{{\includegraphics[width=3cm]{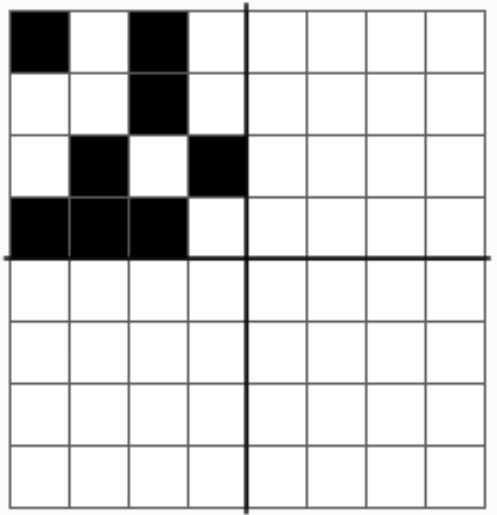}}\label{worst_scenario}}
    \caption{Examples of the best (left) and worst (right) case scenarios for distribution and propagation phases. Cells marked black denote non-zero values and white denote zeros.}
\end{figure}

Somewhere in-between the two extremes described, there is a choice about how to spill non-zero elements from buckets which become too full (due to non-uniform distribution of sparsity pattern) to other buckets. When choosing a bucket to spill elements to, we can define a concept of distance based on the nested iteration (from innermost to outermost around the partitions of $n$, $k$ and $m$) and attempt to minimise this distance for all elements.

\section{CUDA API that were not considered in the current work}\label{AppendixB}
We note that the CUDA libraries offer support for additional, more restrictive sparsity patterns. We did not benchmark these, as there are no equivalent implementations on the IPU, and it is hard to quantify the modelling cost of a stricter sparsity constraint. Therefore, we only seek to compare identical computations on IPU and GPU. The CUDA APIs that we discarded due to this constraint were: 

cusparseSpmm with cuSPARSE blocked ELL format. This format is a block-sparse matrix format with a maximum non-zero blocks per row. It stores an array of block column indices with a marker (-1) for missing blocks. Note that this is one of the few CUDA SpMM formats that supports FP16 compute. It would be possible to use this API to implement a static block-sparse pattern by padding rows to the maximum number of non-zero blocks, but this would presumably have a performance impact.

cuSPARSELt\textsuperscript{\texttrademark} with N:M structured sparsity. This enforces a regular sparsity rule, for example 2:4, which specifies that exactly 2 of every chunk of 4 contiguous weights are non-zero. This is a strong regularity constraint on sparsity patterns and is hard to compare to block sparsity. This method has explicit hardware support in NVIDIA A100 GPUs.

\section{Overview of the static sparse and dense results on IPU}\label{AppendixC}
Figure~\ref{fig:ipu-static-grid} shows a more complete view of the speedup ratio of static sparse to dense, as density $d$, block size $b$, feature size $m=k$ and batch size $n$ all vary. Dark grey blocks indicate missing data (could not fit on single IPU memory). Same conclusions can be drawn as those stated in Section~\ref{sparse-dense-ipu}.
As GPU dense performs similarly to IPU dense (see Figure~\ref{fig:dense-ipu-gpu}), this figure can also be used as an approximate guide to compare IPU sparsity with GPU.

\begin{figure}[!h]
    \centering
    \subfloat{\includegraphics[width=12cm]{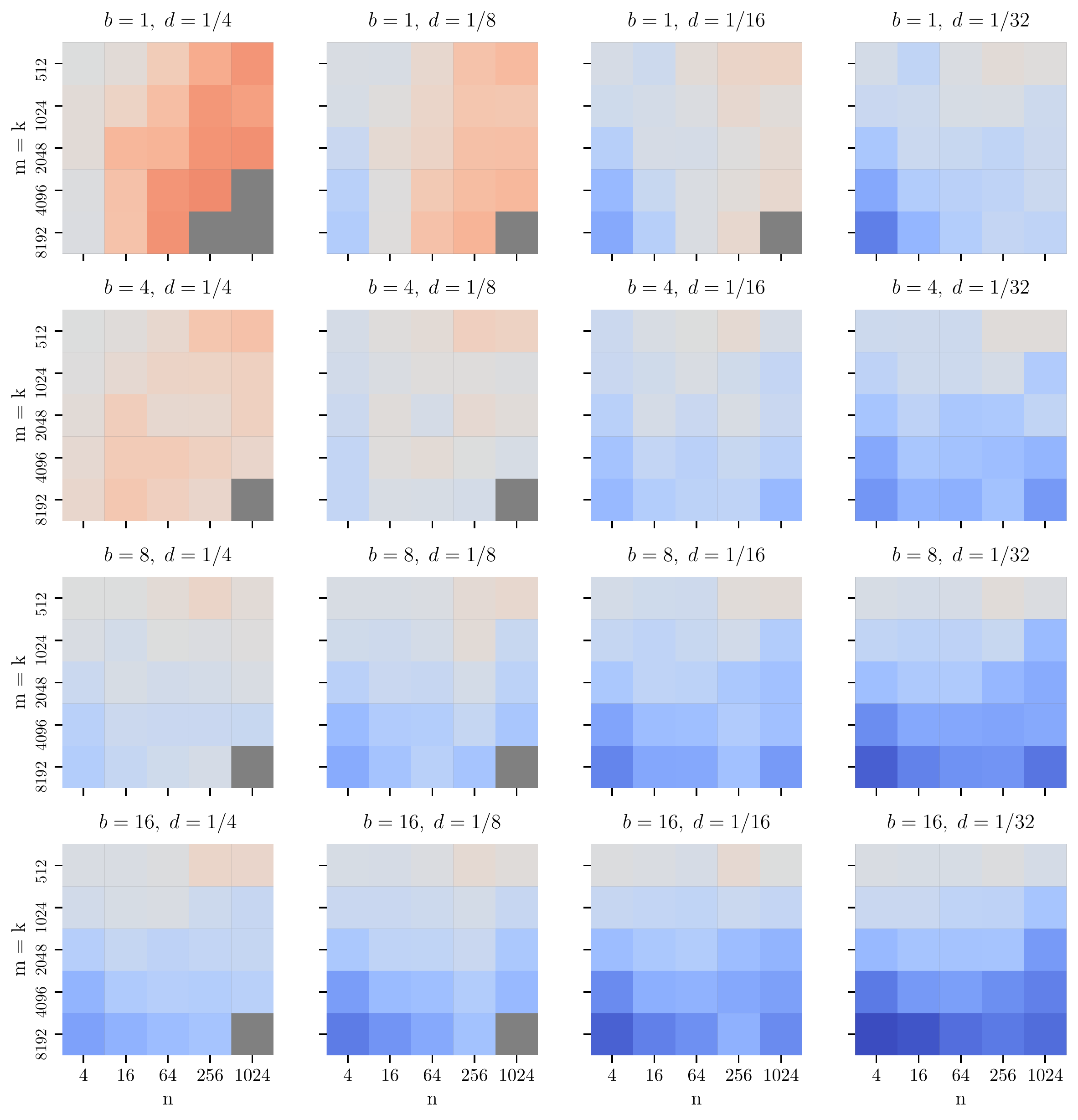}}
    \enspace
    \subfloat{\raisebox{4cm}{\includegraphics[width=1.5cm]{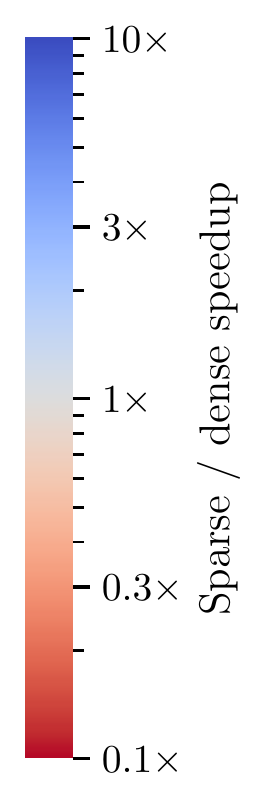}}}
    \caption{Grid showing the speedup ratio of static sparse to dense.}
    \label{fig:ipu-static-grid}
\end{figure}

\end{document}